# A Deep Learning Based Automatic Defect Analysis Framework for In-situ TEM Ion Irradiations

Mingren Shen[1], Guanzhao Li[1], Dongxia Wu[3], Yudai Yaguchi[5], Jack Haley[4], Kevin G. Field[2,6], Dane Morgan[1]

[1] Department of Materials Science and Engineering, University of Wisconsin-Madison, Madison, Wisconsin, 53706, USA

[2] Oak Ridge National Laboratory, Oak Ridge, Tennessee, 37830, USA

[3] Department of Mathematics, University of Wisconsin-Madison, Madison, Wisconsin, 53706, USA

[4] Department of Materials, University of Oxford, Oxford, OX1 3PH, UK

[5] Department of Computer Sciences, University of Wisconsin–Madison, Madison, Wisconsin, 53706, USA

[6] Now at Nuclear Engineering and Radiological Sciences, University of Michigan - Ann Arbor, Michigan, 48109 USA

**Abstract**

Videos captured using Transmission Electron Microscopy (TEM) can encode details regarding the morphological and temporal evolution of a material by taking snapshots of the microstructure sequentially. However, manual analysis of such video is tedious, error-prone, unreliable, and prohibitively time-consuming if one wishes to analyze a significant fraction of frames for even videos of modest length. In this work, we developed an automated TEM video analysis system for microstructural features based on the advanced object detection model called YOLO and tested the system on an in-situ ion irradiation TEM video of dislocation loops formed in a FeCrAl alloy. The system provides analysis of features observed in TEM including both static and dynamic properties using the YOLO-based defect detection module coupled to a geometry analysis module and a dynamic tracking module. Results show that the system can achieve human comparable performance with an F1 score of 0.89 for fast, consistent, and scalable frame-level defect analysis. This result is obtained on a real but exceptionally clean and stable data set and more challenging data sets may not achieve this performance. The dynamic tracking also enabled evaluation of individual defect evolution like per defect growth rate at a fidelity never before achieved using common human analysis methods. Our work shows that automatically detecting and tracking interesting microstructures and properties contained in TEM videos is viable and opens new doors for evaluating materials dynamics.

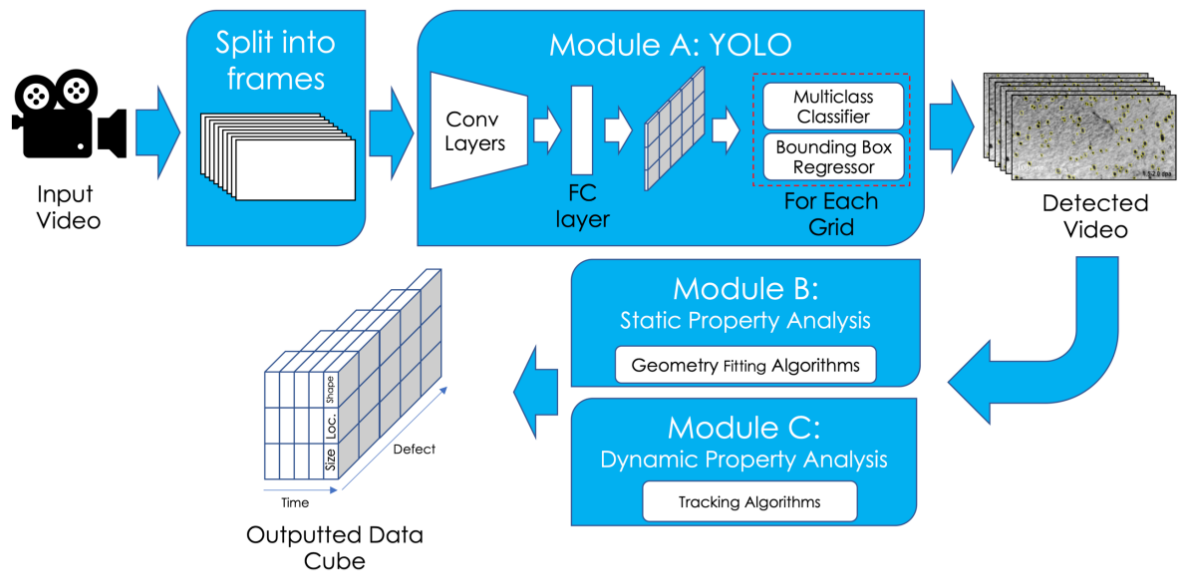

Graphical Abstract – Overview of the developed framework for automated analysis of defects based on the YOLO deep learning method.

**Introduction**

Transmission Electron Microscope (TEM) has widely been used to characterize a material or material system since TEM provides resolution limits at or below common microstructural features of interest. Recently, a surge in the use of in-situ TEM techniques has occurred, partially due to the advent of digital capture devices. In-situ TEM experiments have a distinct advantage over ex-situ experiments as they allow researchers to study materials' intrinsic properties and responses as external conditions are manipulated such as temperature, pressure, and type of reaction cells[1]. In material science, in-situ TEM is frequently used to shed light on challenging problems like elucidating mechanisms for catalysis, atomic behavior during material reactions, and nanoscale property changes under loads[1,2].

The value extracted from an in-situ TEM experiment requires careful analysis of the observed processes. For many of these experiments, this analysis includes dynamically detecting features present in the microstructure and analyzing the microstructural evolution in each frame of the experiment, typically captured in a digital video form. For decades, quantification of defects in in-situ TEM data has been completed by humans, which is tedious, time-intensive, error-prone, biased, and impractical to scale. For example, a typical manual workflow of counting defects in TEM images requires an experienced researcher carefully going through every frame for different types of defects and labeling objects in the images one by one. Such manual analysis typically takes many minutes per frame (e.g., in this study, we found it takes about 20 to 60 minutes to process 1 frame, depending on the complexity of the TEM images). Typical in-situ TEM experiments can generate tens to hundreds of frames of video data per minute, so a long video can rapidly become impractical to analyze. Moreover, the labeling quality also depends on the attentiveness of researchers, which may be reduced after spending hours on this repetitive work. Furthermore, other factors such as researchers' proficiency and personal preference when analyzing TEM images contribute to inaccurate or at least inconsistent labeling. Human interpretations are often required for analyzing TEM images that the same researcher may give different labeling results at different times. The above observations imply that manual counting and analyzing methods are hard to scale and prone to human-based errors. In the future, the demand for better TEM analysis methods will only grow, as recent advances in TEM equipment, e.g., high-speed cameras, and fast microprocessors will keep accelerating the rate of data acquisition[3].

Automatic analysis of TEM/STEM data, especially identifying microstructural defects, is a long-standing pursuit of both the academic community and the industrial sector. To automatically analyze defects contained in TEM/STEM images, various methods have been applied, such as matching key-points in different regions of interest[4], applying different thresholding values to segment different defects[5,6], representing the texture of various targeted structures by the bag of visual words (BoW)[7] or synthesizing artificial image dataset[8], and using traditional machine learning methods, e.g., k-means clustering, to find defects contained in TEM/STEM images[3]. To the best of our knowledge, these methods are only semi-automatic in that they still require extensive human knowledge and time to apply to a given system, and each new material system requires a significant new investment to find an effective approach. Recently, modern deep learning methods have been applied to solve the defect identification problem in static TEM/STEM image data[9–11], suggesting this is a promising approach that could be reliable and highly flexible across many materials and problems. However, deep learning approaches have not yet been applied to the problem of automatic detection and analysis of defects contained in in-situ TEM video. Imaging-related research around TEM/STEM video processing is very active, but has focused on other areas, such as structure reconstruction[12], image quality improvement[13,14], and video alignment[15].

Here, we develop a framework to solve the defect detection and tracking problem common in many materials based in-situ TEM experiments. The framework is centered on the deep learning based YOLO model[16] which can extract, detect, and identify feature populations present in a video to show the evolution history of defects in-situ. Deep learning typically uses a combination of multiple layers of nodes called neural networks to extract the intrinsic structure of input data to build a mapping between the intrinsic structure and targeted output[17]. With advancements in GPU computing powers, accumulation of carefully labeled large scale datasets, e.g. ImageNet[18], and better optimization algorithms like backpropagation, deep learning-based models have shown great success in different tasks such as automatic driving, speech synthesis, and image classifications[17], even outperforming humans in many tasks such as the board game Go[19]. Deep learning has also been widely used in material science[20] and achieves good performances including predicting properties of materials[21,22], identifying material phase transitions[23], and automating the analysis of TEM/STEM data[9–11,24–26]. Among those advancements, defects detection or tracking in TEM/STEM data has yielded human-level or even better than human-level performance, including detection or analysis of dislocation loops, line dislocations, precipitates, and cavities[9–11,24,27–29]. Here we focus on applying deep learning methods and tools for defect detection and tracking in microscopy and we organize the previous approaches to this problem into three different categories. The first one is combining deep learning methods, computer vision knowledge, and other than deep neural network types of machine learning tools together to build the defect analysis system. For example, in Li et al.'s work[9], computer vision methods like Local Binary Patterns (LBP)[30] descriptor were used to describe local pixel environment, machine learning methods like AdaBoost[31] were used to select the most useful visual features, and shallow Convolutional Neural Networks (CNNs)[32] were used to refine the final output to find the defects. This type of methods requires extensive hand-tuning and integration of multiple stages which is unlikely to provide a practical general approach compared to relying on just one deep learning model. The second main type of methods belongs to the encoder-decoder type of deep learning methods[33]. For example, Ziatdinov et al. used a weakly supervised encoder-decoder type neural network system to extract atomic locations and defect types from atomically resolved scanning transmission electron microscopy images to interpret complex atomic and defect transformation of silicon dopants in graphene as a function of time[26]. One special type of encoder-decoder type of methods called U-Net is also popular and widely used for analyzing material images like studying segmentation of nanoparticles in bright field TEM images[34] and defects in STEM images of steels[11]. An encoder-decoder model extracts the most relevant information and builds a useful inner state, which usually requires careful training[33]. The third type of methods relies on using mature object detection methods like Faster R-CNN[35], YOLO[16], Mask R-CNN[36], and SSD[37] etc. For example, Mask R-CNN is used by Chen et al. to study the microstructure of aluminum alloy[38] and Faster R-CNN is utilized by Anderson et al. to investigate helium bubbles in X-750 alloy under irradiation[27]. These methods tend to be quite accurate and fairly easy to train. Given their wide use and appealing properties, in this study, we use one of the third type of methods (YOLO) to study geometrical and temporal changes of loop type defects in the videos of In-situ TEM ion irradiations. YOLO is similar to methods used by others for TEM but unique in its speed to analyze videos in real time.

One of the most successful applications of deep learning is computer vision, where the ultimate goal is teaching a computer to do the image-related task(s) like finding which object contained in an image (object detection) and which pixel belongs to different objects (image segmentation)[39]. Since a breakthrough in the ImageNet Large Scale Visual Recognition Challenge (ILSVRC) was made in 2012, the deep Convolutional Neural Network (CNN) based approach has demonstrated its success in many image-related tasks[39]. For the object detection problem (trying to find the location and category of all the objects contained in an image), which is also the focus of this project, there are two general categories of methods: two-stage methods and one-stage methods[39]. For a two-stage method like Faster R-CNN[35], the object detector will first propose some candidate bounding boxes containing the object location information and then classify the category of those candidate bounding boxes. One-stage methods like YOLO will output the object location and category at the same time[16]. Typically, two-stage methods are slower but more accurate than one-stage methods. YOLO (which is an acronym for You Only Look Once), is one of the most widely used one-stage methods and offers speed, accuracy, and fast engineering application potentials[39]. The key ideas of YOLO are dividing the whole image (or video frames) into grids and predicting the location and the category of the potential bounding boxes with a set of pre-defined anchor boxes in each cell of the grid[16]. YOLO keeps improving its design and implementation details over time and during the writing of this draft, YOLOv4[40] and YOLOv5 (https://github.com/ultralytics/yolov5) were developed. However, we will use YOLOv3 as this is still the most widely used and recognized version[16,41–43].

In this paper, we focus on the specific task of adapting the deep learning based YOLOv3 model into an automated framework for analyzing in-situ TEM video data. Specifically, we focus on the problem of detection and analysis of radiation-induced dislocation loops generated by an in-situ ion irradiation TEM device, the Intermediate Voltage Electron Microscope (IVEM) housed at Argonne National Laboratory (ANL). This in-situ ion irradiation TEM device introduces controlled ion beams into a TEM to achieve a high number of atomic displacements per atom (dpa) to mimic the irradiation environment a material will experience in nuclear reactors, satellites, or space stations[44]. The device enables co-irradiation and observation of radiation-induced defects using diffraction-based contrast while the material is being irradiated.

**Material and methods**

The in-situ ion irradiation TEM video-based data used within this study has been previously studied and analyzed using the common typical human analysis method[44]. Extensive details on experimental design, human analysis, and resulting materials effects have been previously published[44]. Here, we only present the most pertinent details for context. We selected one of the four model samples (Fe-18Cr-3Al) from the previous study for the current study, but the YOLO-based methods can be easily generalized to other samples or different material systems. The Fe-18Cr-3Al in-situ ion irradiation TEM video-based dataset was generated by performing in-situ irradiation using the IVEM-Tandem Facility at ANL with a pre-thinned TEM specimen titled to the **g**=011 strong two-beam conditions with a frame rate of 15 frames per second using a Gatan 622 video camera. The irradiation was performed using 1 MeV Kr$^{++}$ ions up to 2.5 dpa at a constant temperature of 320°C and a dose rate of 8.3x10$^{-4}$ dpa/s. Note, dpa is a measure of the damage or energy imparted into the system and it is known that ion bombardment at the dpa ranges observed generate embrittling defects in Fe-based alloys[45,46]. Under these radiation conditions, it was expected that two dislocation loop types would form, one with a Burgers vector of a/2⟨111⟩ and the other with a⟨100⟩ [44]. Under the strong bright-field two-beam condition used where **g**=011 and the deviation parameter, $s_g$, close to zero, only a fraction of a/2⟨111⟩ and a⟨100⟩ loops are visible in TEM. To enable direct comparison to the previous human-based analysis where multiple **g**-vectors were used for detection and analysis[47,48], we applied a fractional visibility constant 7/4 to make YOLO detection results comparable with published results using other **g** conditions[44]. Defect size was estimated by assuming the defects are elliptical and the defect size is the length of major axes of the ellipse. The video image size is 1344 pixel x 962 pixel with 2.6884 pixel/nm conversion factor which gives the physical size as 500.0 nm x 357.8 nm. The video consisted of 1175 frames which were linearly related to dpa and time through **Equation 1** which means each frame corresponds to 1.75 seconds.

$$Time\ (s) = \frac{dpa}{8 \times 10^{-4}}$$

$$dpa = 0.8534 + \frac{[(Frame\ Number) \times 1.6466]}{1175} = 0.8534 + (Frame\ Number) \times 0.00140$$

**Equation 1**

The video data was acquired via frame-by-frame image registration to eliminate sample drift and relevant camera movement. Since there is no landmark frame or feature that can be used to align the video across the whole irradiation dose range, the video was divided into smaller batches for primary image registration, and the final sets from the previous batch were used to carry over the alignment[15,49]. The alignment was done by frame registration based on the selected landmark frame with a template matching and slice alignment plugin[50]. The dissemination of data and codes for this paper is described in the Data and Code Dissemination section.

We opted not to use any previous data labeling and decided to label the data ourselves for this project to establish the ground truth data. This choice was because we did not have the exact pixel positions of each defect in the previous study by Haley, et al[44]. We followed the labeling process that has been used in other studies[9]. The ground truth data was labeled by two trained researchers and they checked each other's labeling and explanations for 3 frames before labeling the real data via an open-source software called ImageJ[51]. Their labeling will be treated as ground truth in this study.

The YOLOv3 model was adopted from an open GitHub repository (https://github.com/qqwweee/keras-yolo3). To train the model, we first converted the pretrained darknet53 weight via COCO dataset into Keras format and then modified the final class number to our defect number, which was one class in our case since we treated all a/2⟨111⟩ and a⟨100⟩ loops as the same type of defect. This single class approach was necessary as Burgers vector determination is not possible using only the **g**=011 condition in the video. We then applied the transfer learning technique to fine-tune the model by freezing the first 245 layers of YOLOv3 and training the last 7 layers[52]. The in-situ ion irradiation TEM video data in this study was composed of 1176 frames and 21 frames were selected and labeled. The sampling was done at random except an effort was made to assure that the sampled frames were approximately uniformly distributed throughout the full set. Among the sampled 21 frames, 15 frames were used for training (trained on 12 frames and validated on 3 frames) and 6 frames were used for testing, where these 6 frames were not seen by the YOLO model during training. The model was trained on GeForce GTX 1080 for 18300 epoch and the learning rate of Adam optimizer was switched between $10^{-4}$ and $10^{-5}$ with batch size equal to 4 and Non-Max Suppression (NMS) IoU equals 0.45 to find the optimal weights. Real-time data augmentation operations, e.g., left-right flip, changing hue, saturation, lightness, were applied for the training dataset to enrich the dataset and enhance the performance of the CNN for variations in defect contrast, size, and morphology in the video data set. Real-time augmentation works by augmenting at each training epoch, generating new augmented images in each epoch.

**Results and Discussion**

We first tested the performance of the trained YOLO model qualitatively by comparing the detection result of testing frames to that of the ground truth labeling visually, as shown in Figure 1. In general, the automated machine learning program labeled results agreed with the ground-truth labeling by humans, except for certain ambiguous grey spots and when there existed several touching adjacent loops. A zoomed-in comparison between the ground truth labeling and YOLO predictions is shown in Figure 2. The model was also run on the whole in-situ ion irradiation TEM video. In general, the YOLO model successfully detected nearly all the dislocation loops. The original TEM video and YOLO prediction overlaid are provided as described in the Data and Code Dissemination section.

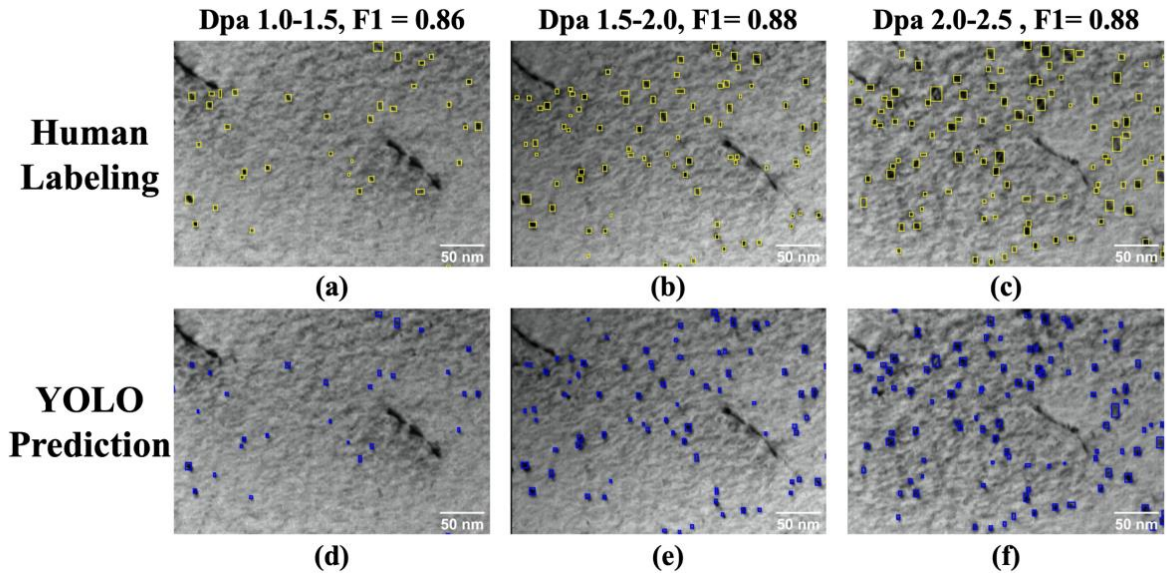

**Figure 1. Selected images from the test dataset for various damage doses (e.g. time scale). Subfigure (a), (b), and (c) are the ground truth labeling developed by two researchers, while subfigure (d), (e), and (f) are labeled by the automated machine learning program. Here, (a) and (d) are for frame number 120, (b) and (e) are for frame number 472, and (c) and (d) are for frame number 824. 1 frame increment equates to about 0.00140 dpa, see Equation *1*. F1 score compares the machine detection results with human labeling of each column separately.**

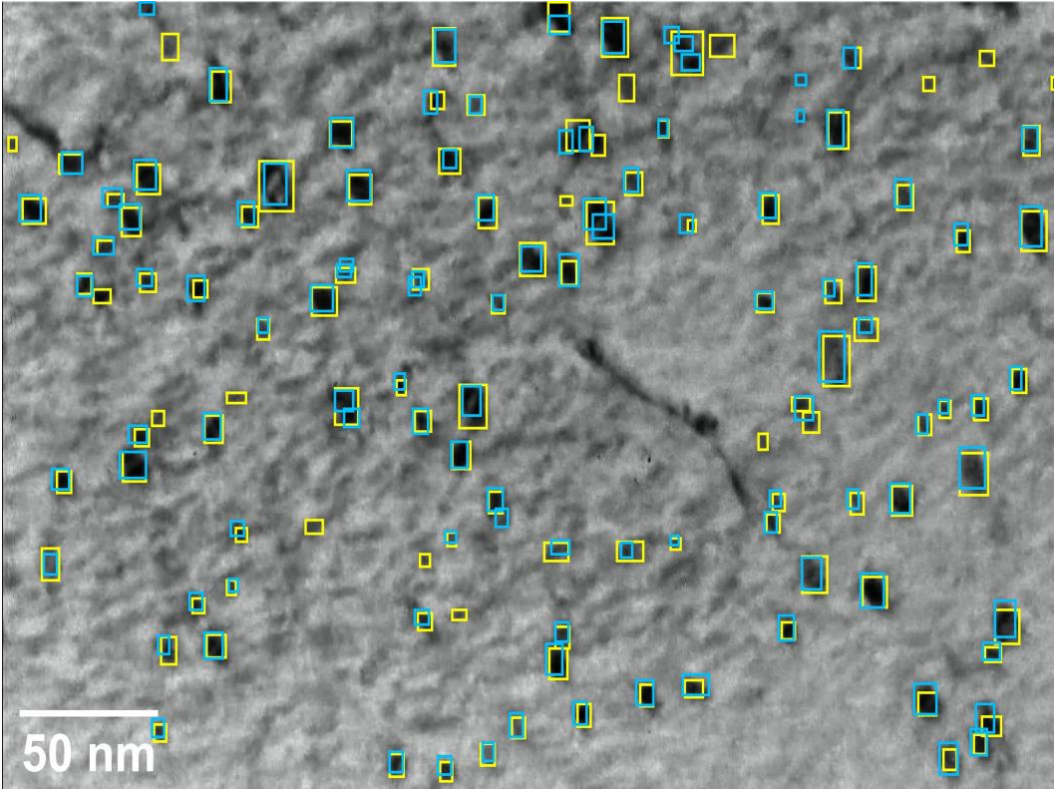

**Figure 2. The visualized comparison of the human labeling results (blue boxes) to the YOLO detector results (yellow boxes) for frame number 824.**

The initial qualitative comparison was encouraging so additional quantitative analysis was conducted. The statistics of the model performance were examined based on the metrics of precision and recall and their harmonic mean which is also called the F1 score. The precision, recall, and F1 score were generated using the six test images that were never used in the training process. The test was iterated with different cut-off Intersection-over-Union (cut-off IoU) values as shown in Figure 3. Here and elsewhere in the paper the IoU refers to the ratio of the area of overlap (intersection) to the combined areas (union) of predicted and ground truth bounding boxes. The cut-off IoU refers to the threshold above which a predicted bounding box is considered as a candidate match for a ground truth bounding box. Predicted bounding box matches are assigned by building a matrix of all IoU values between all predicted and ground truth bounding boxes and making assignments between predicted and ground truth defects using the highest IoU in the whole matrix. When an assignment is made all the matrix entries associated with those predicted and ground truth bounding boxes are removed from the matrix and the process is repeated. This approach provides a unique assignment and effectively assigns the highest overlapping predictions to the appropriate ground truth boxes. In general, a lower cut-off IoU means higher tolerance on the discrepancy between the machine labeled region and the human-labeled region, which agrees with the trend shown in Figure 3, indicating that the performance of the trained model increased as the cut-off IoU decreased.

We selected a cut-off IoU = 0.15 to assess the performance of our model. This value is lower than usually used in machine learning classification problems, but we believe is reasonable for the following reasons. Many defects are small so a shift of just a few pixels in the size and/or center of the ellipse can lead to significantly reduced overlap in bounding boxes. Such shifts are likely within the realm of the uncertainty of human labelers, and of course, the YOLO algorithm makes some location errors, so relatively small cut-off IoU can occur even when two bounding boxes are clearly finding the same defect ellipse. Furthermore, from the density calculation showed below, defects are typically much farther away than their size, with a typical distance at the 2.5 dpa (where defect density is $3 \times 10^{16}$ cm$^{-3}$) of about $\frac{1}{\sqrt[3]{3 \times 10^{16} \text{cm}^{-3}}} = 32$ nm. This separation length scale makes it unlikely that boxes of sizes ~ 6-10 nm (the median defect size) on a side will be assigned to the wrong defect just due to allowing a modest overlap.

The F1 score obtained for the cut-off IoU = 0.15 is very encouraging as scores in the range of 0.85 to 0.95 are typically considered very good for object detection results[43,53]. Furthermore, at this cut-off IoU, our 6 testing images are all reasonably accurately modeled, with F1 scores ranging from 0.83 to 0.93.

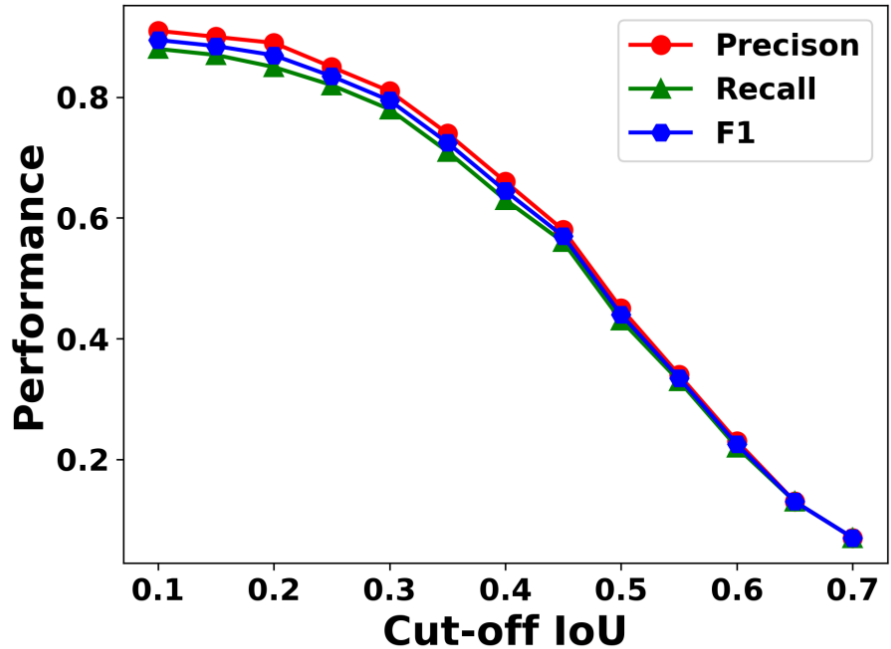

**Figure 3.The performance of the YOLO detector with different cut-off IoU thresholds.**

The developed YOLO model was run on each frame of the in-situ TEM video to extract geometry information of each visible defect for the duration of the experiment. After obtaining the geometry and position of each defect per frame, we used this information to extract defect properties, such as median size and number density. Such properties of materials are widely used in the nuclear materials field from which the dataset originated and provide insights into the interplay between the imparted damage and the change in microstructure. We picked four typical frames and compared the machine learning prediction results with ground-truth labeling. Those 4 frames were not used in training and testing the machine learning model. We first compared defect density. Defect density is important for many materials properties and for nuclear materials as it is strongly correlated to mechanical properties, e.g., through the dispersed barrier model[46,54]. Loop density comparisons between machine learning results and our labeling results are summarized in Figure 4. The densities of defects per frame were determined via machine learning (ML) method and manual labeling for ground truth by dividing the total number of the loops by the volume of the sample for each frame. The sample was treated as a rectangle bulk with dimensions of 416.6 x 264 x 75 nm$^3$ and both results were corrected based on the loop invisibility for the given imaging condition.

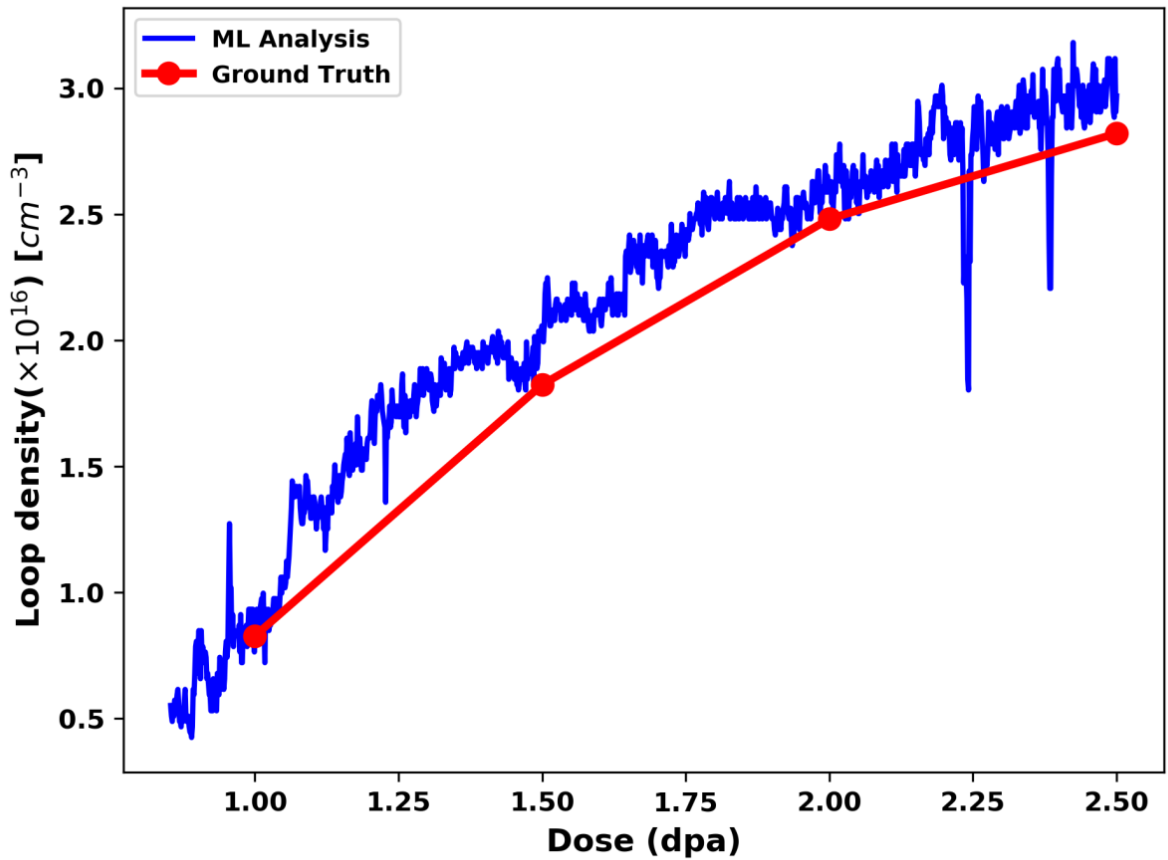

Figure 4. Loop number density from the whole TEM video. The plot compares the loop number density obtained from the ground truth labeling done by experts in this study and the result obtained from the machine learning detector. All the data shown on the plot uses the corrected proposed density, which is 7/4 of **the raw density (see Sec. Material and methods). The sudden drops in the late stage are an artifact arising from camera motion.**

Both techniques for analyzing the in-situ videos showed a general trend of increasing loop density with irradiation dose (time) which was expected based on general radiation effect theory and previous analysis of the experiment[55]. Overall, machine learning results were close to the ground truth labeling results throughout all frames, varying at most 12% compared with ground truth labeling at the four measured points data in Figure 4. We believe that the observed discrepancy between machine learning and ground truth data in Figure 4 is likely comparable to different researchers' preferences in labeling ambiguous loops and perhaps cannot be significantly improved without more consistent labeling. It is noteworthy that the sudden drops observed in the late stages arise from abrupt stage movements that rapidly alter the field-of-view and momentarily artificially reduce the effective number of loops observed. This effect is similar to camera movements in the traditional sense.

After obtaining the density of defects from the machine learning detector, we then used a watershed fitting method to determine the morphology statistics of defects. Since all images were recorded with metadata to allow for pixel to physical distance conversions, we could predict the geometric information of each detected defect based on the pixels involved in the defect. We used the watershed algorithm provided within OpenCV[56] to determine the defect pixels and their boundary. Watershed is a commonly used image segmentation method, which divides different objects with watershed lines and then, based on the contour found, extracts precise information about the defects' position, size, and orientations[57]. We used OpenCV's marker-based watershed algorithm. This method requires users to initially label pixels according to their belonging to one of two categories, referred to as the "sure object" and "sure background". The sure objects and background were found by applying a thresholding method, specifically Otsu's binarization and Distance Transformations. To remove noise, we use a morphological opening operation with a 3x3 kernel. We followed the official tutorial from OpenCV, and more details can be found there[58]. Watershed found boundaries of defects and backgrounds, but the boundaries were not very smooth. OpenCV's `fitEllipse()` function was called to fit the needed defects and the major axis length of the fitted ellipse was defined as the defect size. Detailed fitting results with a cropped region of interests are provided in the Data and Code Dissemination section. The machine learning results of the defect size distributions were compared to ground truth labeling in Figure 5. Although differences were observed in defect median size of each frame in Figure 5, investigation indicated that these differences did not exceed 13.0% difference in median size, and the average difference is only 5.5% and the standard deviation of difference is 5.3% across all doses investigated. The exact formula used to calculate these statistics is given in the Supplemental Information (SI) Section 1. These results indicated that a well-trained machine learning based model could be used for loop detection and analysis and achieve human-like performance comparable or better than the large differences that can be expected by manual labelling[9]. The boxplot comparison provided in Figure 5 showed the viability of the machine learning results. At the same time, it needs to be emphasized again that the true strength of such a technique lies in its ability to detect defect information for every frame quickly and accurately instead of just focusing on a small subset of frames.

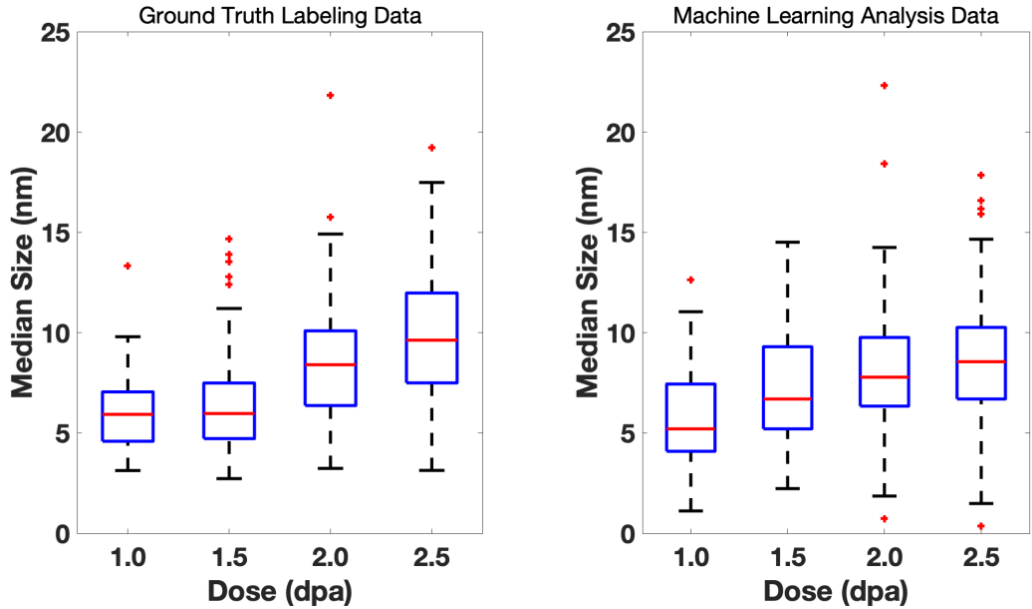

**Figure 5. Box plot comparing the distribution of median size for two methods the ground truth labeling done by experts in this study, and the result from machine learning detector. All distributions are separately analyzed and compared by their irradiation condition, which is 1.0, 1.5, 2.0, and 2.5 dpa.**

Figure 6 shows the size distribution for the entire duration of the in-situ ion irradiation TEM experiment where, for each frame, the blue line represents the median of loop size, the top of the gray boundary indicates the third quartile of loop size distribution and the bottom of the gray boundary indicates the first quartile of loop size distribution. With the YOLO-based machine learning detector, we could extract data generated in every frame and investigate the material properties with hundreds of times more data than previously collected by hand for this data set (the data collected by hand are the red points shown for four different typical dpa values where these four points are not seen in training data, with red lines connecting them as a guide to the eye). The large amount of analyzed data makes subtle trends easy to identify. For example, although there are some noises, a clear trend can be seen in Figure 6 that the median size, Q1, and Q3 increased as the dose value increased from 0.83 to 2.3 dpa and remained stable from 2.3 to 2.5 dpa. Such a result agreed with the relationship found in Haley et al.[44].

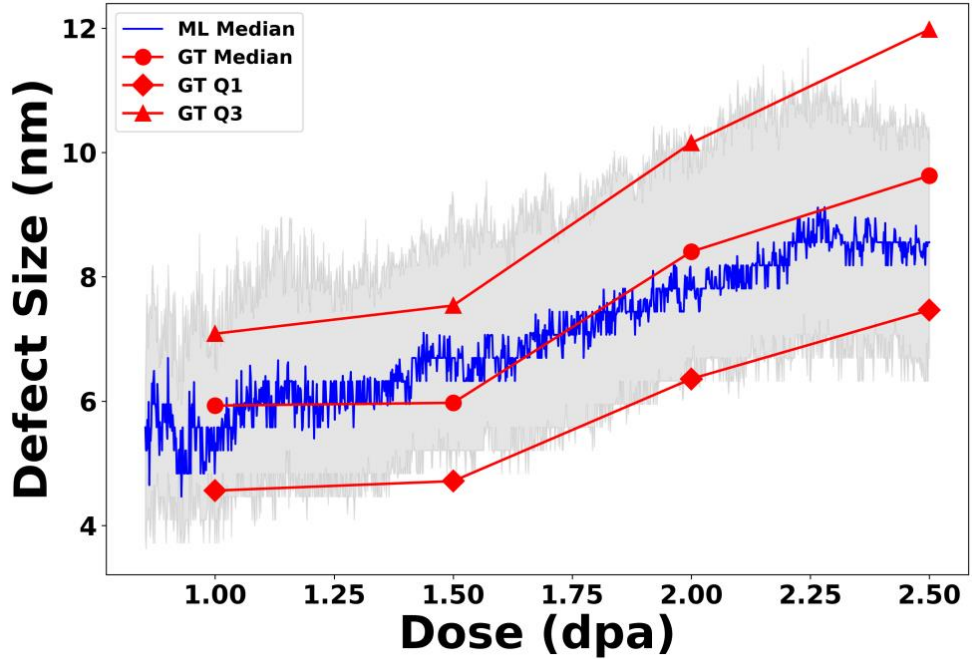

**Figure 6. Change in size distribution as a function of irradiation dose based on machine learning detection. It can be found that the median size (Blue line), first quartile Q1 (Upper gray boundary), and third quartile Q3 (Lower gray boundary) increase as the dose increases when the dose is from 0.83 to 2.3 dpa. Median size, Q1, and Q3 stabilize above ~2.3 dpa. Red lines connect the red points that represent the ground truth labeling of median size (circle), first quartile Q1 (triangle), and third quartile Q3(diamond) of 4 typical frames to provide a guide to the eye.**

One of the most exciting applications enabled by automated data analysis of in-situ TEM data is the ability to track all defects as a function of time (i.e., frame). With this application in mind, we developed a tracking module based on YOLO output to track defect motion in the data. Since video is sequential images in time, we can track defects by counting and measuring their sizes across frames to discover their evolution in morphology and mobility under irradiation.

This process is usually called object tracking in computer vision studies and is important for applications such as surveillance and security systems, traffic monitoring, human-computer interaction, etc[59]. One of the most widely used methods for object tracking is tracking-by-detection, also called tracking-by-repeated-recognition[59,60]. In this method, tracking is achieved by detecting targets in consecutive image frames with trained object detectors and linking detected objects across frames to generate the tracking results, e.g., trajectory or motion data[60]. We used Trackpy, a Python package for particle tracking, to link the detected objects generated by the machine learning detector. Trackpy implemented the algorithms first developed for colloidal particles by John C. Crocker and David G. Grier[61] in Interactive Data Language (IDL) and the algorithm worked well for both non-interacting and interacting systems[62]. Trackpy is widely used in the soft matter community for tracking the movement of particle-like objects e.g., colloidal particles or cells in microscopy videos or images. A typical workflow of Trackpy can be split into three steps: (1) Locating Particles, (2) Refining Location Estimates, and (3) Linking Locations into Trajectories[61]. In the first step general features of particles like diameter, maximum size, and separations are used to locate all peaks of brightness in the image which includes the initial object coordinates. Subsequently, more pixel-level information is used to distinguish real particles from spurious ones. Finally, the locations of particles in each image are matched with corresponding locations in later images to yield the whole trajectories. The tracking module is a powerful tool to obtain several important statistics relating to the motion and evolution of defects. When combined with automatic labeling, it provides a new way to study defect dynamics under irradiation at a fidelity not possible using previous methods. We demonstrated this advantage by two case-studies using the tracking algorithms: (i) studying defect evolution and trajectory of interesting defects and (ii) extracting statistics of individual defect mobility e.g., diffusion coefficients.

To study defect evolution and trajectory of interesting defects, we first showed the size change of an individual defect, then compared the trajectories of the slowest moving defect and the fastest moving defects, and finally, showed the landscape of defect moving trajectories. In Figure 7, a defect was shown to undergo significant size change as the dose increased. With the help of the tracking module based on the YOLO and Trackpy package, a full history description of a defect's size change was recorded to illustrate the evolution of the defect. The defect size change is shown in Figure 8 which clearly indicated that defect size increased as radiation dose (dpa) increased.

Figure 7 and Figure 8 show the ability to extract a single defect growth evolution as part of this in-situ TEM experiment. It is interesting to note that the shape/trend of the growth rate for individual defects varied, with some showing unconstrained linear growth and others showing asymptotic growth, and even some showing growth followed by shrinkage. Although not the focus of this study, we believe the different growth curves for individual defects could be attributed to local variation in the direct vicinity of the defect, and these variations could promote or retard growth under irradiation. Significantly more analysis of the data would be required to evaluate the postulated mechanism. But even at the level of the analysis presented here, the power of such individual defect tracking is obvious.

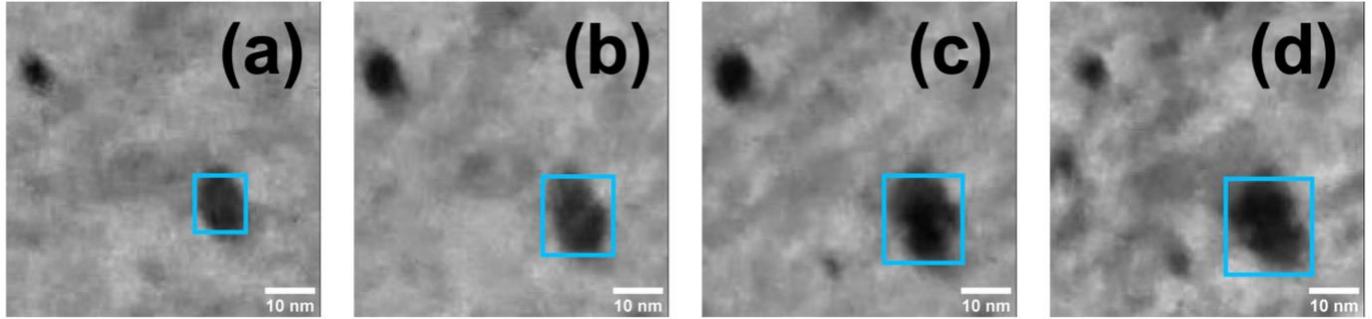

**Figure 7. Reduced field-of-view bright-field TEM images of a single dislocation loop growing under increasing irradiation dose of 1.28 displacements per atom (dpa), 1.72 dpa, 1.95 dpa, and 2.35 dpa for a)-d) respectively. The highlighted loop shows the dynamic change in contrast necessary for the tracking model to detect and quantify. The defect id (51) was assigned by Trackpy.**

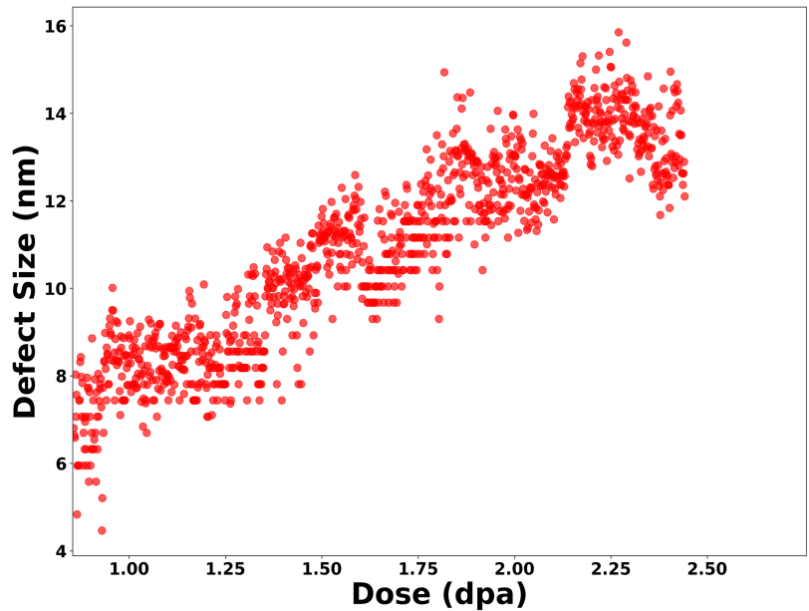

**Figure 8. The size change of a single typical defect, which is the same defect shown in Figure 7.**

Since the individual history of every defect was obtained, it was straightforward to examine defects with interesting behaviors. For example, as shown in Figure 9, our tracking module could determine the motion of very slow-moving (Figure 9(a)) and fast-moving (Figure 9(b)) defects.

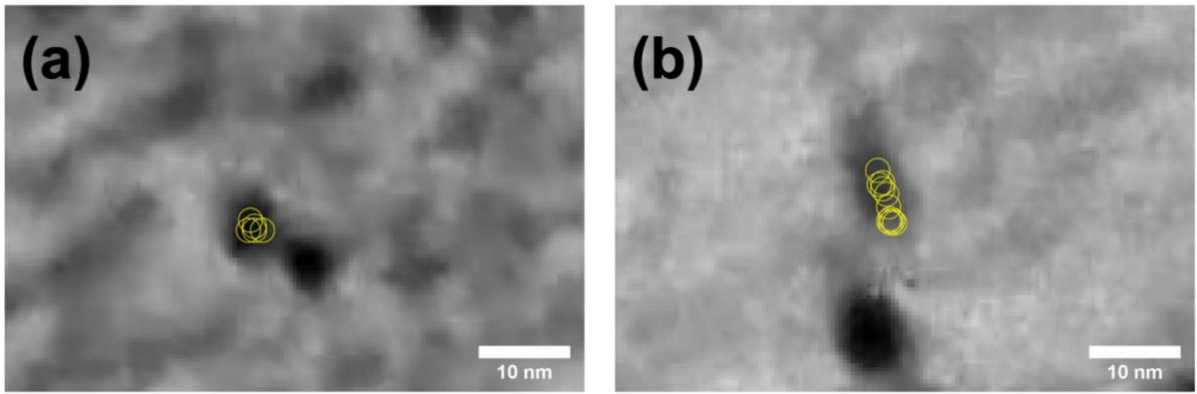

**Figure 9. Trajectories of two typical defects throughout their lifespan. (a) represents a typical defect that has close to minimum diffusion coefficient value. Figure (b) represents the defect that has a nearly maximum diffusion coefficient. Each yellow circle center represents a specific location of the defect in certain frames, and the set of locations are plotted on a single image to show the relative movement. The defect id is assigned by Trackpy.**

The spatial distribution of defect trajectories was also an interesting property that was determined and is shown in Figure 10. It is noteworthy that in the original video source, due to thermal expansion of material and TEM user operations under irradiation, the viewable area adjusted somewhat over time. This movement is an artifact of the in-situ experiment but the Trackpy package corrected for these artificial movements enabling us to target only the real movement of each defect.

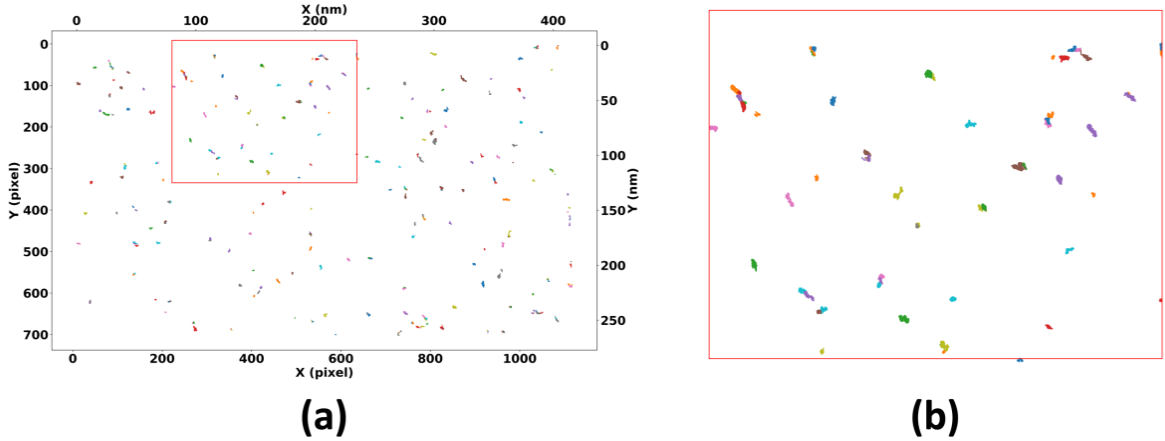

**Figure 10. The trajectory of typical defects detected in TEM video. The movement of this type of defect is roughly cyclic, so the trajectory is not a single line but rather a small group of points. Results were generated by Trackpy. Subfigure (b) is a zoomed-in result of the red rectangle in subfigure (a).**

Since we knew each defect's position and time stamp, an effective two-dimensional diffusion coefficient ($D_{eff}$) can be determined. Diffusion of defects is an important property of defect behavior in nuclear materials[63]. We calculated this effective diffusion coefficient using the following relationship:

$$D_{eff} = \frac{|r(t+\tau) - r(t)|^2}{4*\tau}$$

Note that $D_{eff}$ is not a true diffusion coefficient as we make no effort to correct for the two-dimensional projection of the three-dimensional defect motion, which can be complicated by the exact angle of the sample and the detailed motion of the defect. Our goal for this work is merely to demonstrate the ability to track trajectories through the combination of YOLO and tracking tools, not to perform detailed analysis to extract physically meaningful diffusion coefficients. To perform the analysis, we choose 345 consecutive frames (from 1176 total) over which the camera appears to be very steady. These frames are from frame numbers 461 to 805, corresponding to dpa values from 1.50 to 1.98. Only the regions away from the edges of the figure are used to avoid defects appearing and disappearing due to small changes in the image region. Specifically, we consider only the region with Y position from 200 to 1450 pixels and X position from 250 to 2150 pixels where the original size of the image is 1728 pixels in the Y-axis and 2412 pixels in the X-axis. We find a total of 741 defects in Trackpy, which is significantly larger than the number of defects in a given frame. This larger value is due to the fact that defects have a finite lifetime due to their appearing over time and, in some cases, disappearing, which leads to more tracked defects than actual defects in the analysis. Our average lifetime is 54 frames. While some of the defects may actually appear and disappear, many of these events are clearly artifacts due to Trackpy inadvertently assigning multiple global IDs to the same defect, which effectively causes one defect to disappear and another to appear even when it has not actually done so. Such errors make our defect counts inaccurate from TrackPy but do not lead to incorrect estimates of the defect's $D_{eff}$. To illustrate the values of the diffusion coefficients, in Figure 11 we show the distribution of $D_{eff}$ as a function of binned defect median sizes. We used "median defect size" as a defect would have different sizes in each frame where it is identified, either due to small changes in size estimates from the numerical analysis or due to the defect growing during the irradiation. We then calculated the average $D_{eff}$ of defects that fall into the same bin, where we have 50 bins from 2 nm to 18 nm. While this figure illustrates the type of correlation one can explore with the automated data analysis, in this case we find no statistically meaningful trend with defect size.

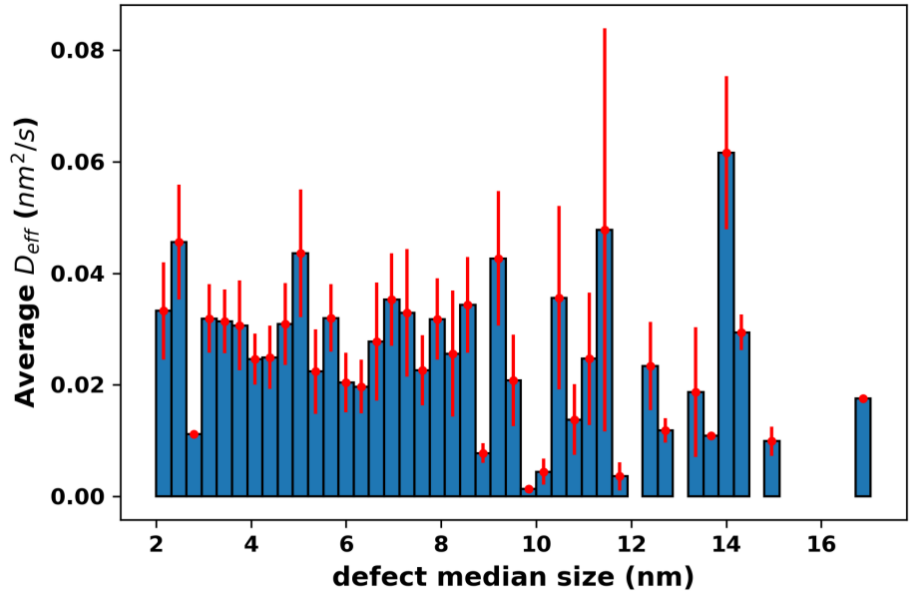

**Figure 11.** The distributions of the effective diffusion coefficient $D_{eff}$ calculated by Trackpy as a function of the defect median defect size. The data is presented as a histogram with each bin of width 0.32 nm, giving 50 bins from 2 nm to 18 nm. The height for each bin is the mean $D_{eff}$ of all defects in that bin. Error bars are the standard deviation of the mean. We use "median defect size" as a defect will have different sizes in each frame where it is identified, either due to small changes in size estimates from the numerical analysis or due to the defect growing during the irradiation. No error bars are given for bins with just one defect as the errors cannot be readily estimated.

## Discussion and Conclusion

To further validate the detection results generated and analyzed by ML methods we developed in this study, we compared our results with those previously completed by Haley et. al. who investigated the same TEM video with conventional manual analysis method[44]. Based on the comparisons, we concluded that the results generated by our ML method are close to those determined by human experts. For example, the discrepancy between the ML generated loop density differs from the results in Haley et al. by at most 38%. And the difference can be largely attributed to the researchers' preferences, as different experts may have different labeling preferences for ambiguous objects. Likewise, the difference between statistics of median size distribution from ML and Haley et al. did not exceed 32% difference in the mean size, 24% difference in median size, and 30% in the standard deviation across all doses investigated. Equations for determining these statistics are given in the SI Section 2.

It is important to be sure that the model is robust to at least some reasonable levels of noise. To test the sensitivity of the model we used scikit-image (https://scikit-image.org/docs/dev/api/skimage.util.html#skimage.util.random_noise) to add Poisson, Salt and Pepper, and Gaussian additive noises to the test images. We calculated the precision, recall, and F1 scores from the model for a range up to quite significant added noises and the impact on the performance is less than 20% in the F1 score for all cases. This impact is relatively minor and suggests our model is quite insensitive to noise. Detailed information is provided in SI Section 3.

Since the YOLO object detection model performance is lower for very small objects[64], there exists a threshold of defect size below which our model cannot detect a defect. Similarly, there is a defect size below which human labelers do not label a feature as a defect. It is important that the human lower limit is larger than the YOLO lower limit or otherwise we will systematically fail to identify very small defects. The human labeling threshold value was estimated as 7.24 pixels (2.69 nm) based on the lower limit in our labeled data. The YOLO object detection algorithm finds defects as small as 1.86 nm, so YOLO is able to find defects as small as any human chooses to label.

Although the performance of the detector we used was quite accurate for defect recognition in TEM video, improvements to the model are needed. Errors likely could be reduced by more extensive optimization. For example, in the training, we only used default anchor box settings and K-means clustering of bounding box sizes in training data could be a better way to find the best set of anchor boxes. Also, more data augmentation operations could be applied e.g, rotation, adding noises, and cropping or affine transformation to achieve better performance. Errors could also be reduced by removing biases and ambiguities in the labeling. For example, it was often unclear how to establish the ground truth labeling of closely distributed objects with no significant white space between two centers.

It should also be noted that the images used in this study are of very high quality, with limited noise and few confounding contributions (e.g., surface oxide), and undergo fairly modest changes during the irradiation (e.g., few defects move significant distances). The high-quality and modest changes of the images almost certainly help the model performance and subsequent defect tracking. Furthermore, we focus our model on only one type of defect, a single category of dislocation loop, to reduce the burden of labeling and focus on the most prevalent defects in the images. Many samples will have other types of defects (and some are even present in our images, e.g., dislocation lines), and tracking these is an important area for future study. While there is nothing intrinsically limiting YOLO to just one defect type (YOLO could be extended to 9000 classes of objects[65]), what we studied in this paper is a very simple case. To fully assess the general effectiveness of our approach and develop a broadly applicable tool, the model needs to be demonstrated on many more data sets with multiple defect types, varying image and sample quality, and more complex defect evolution during irradiation. However, the present deep learning model is a powerful proof of principle and suggests that a broader program may be successful and have a major impact on the defect detection community.

For future study directions, we think two major directions are worthy of investment. One is creating high labeling quality data sets. For example, in this study, we combine a/2⟨111⟩ and a⟨100⟩ loops together to alleviate the labeling burden, but it will be more informative if we can differentiate these two types. Such high-quality labeled data does not necessarily have to come from experiments and synthetic data can have many advantages. For example, image simulation, such as the multi-slice method, can generate high-quality images filled with known types of defects[66]. This method can help avoid the tedious, error-prone labeling process. Synthetic images might also be generated with deep learning methods such as Generative Adversarial Networks (GANs)[67,68], which are powerful tools for generating images similar to an existing set. GAN generation might be done in such a way that labeling is automatic, creating an almost unlimited supply of high-quality labeled synthetic images or converting images collected from different conditions to the condition for which our model is trained, allowing the community to better utilize limited labeled data[69,70]. The second direction worth exploring is to apply the analysis system developed in this paper to TEM devices to provide real-time statistics and even direct labeling of defects (e.g., with a fitted ellipse) in images to guide users during experiments. This approach is similar to the real-time Augmented Reality (AR) methods that have proven to be useful in biological microscopy studies[71]. This combination will provide a straightforward, real-time output of deep learning analyzed results for TEM studies and the material community.

In summary, the present work shows that if the accuracy obtained here can be extended to more general and complex data, these deep learning tools are a potentially transformative methodology for the TEM community. The YOLO based system developed in this study provides an automatic, fast, and reliable quantitative analysis of both position and morphological evolution of defects in frame level. Furthermore, the YOLO based system can help researchers track the motion of defects, which will allow new levels of dynamical analysis. Furthermore, the approach is easy to use and adapt to other sets of experiments. The speed of YOLO means that it can be used in real time to adjust experimental conditions (e.g., dpa, temperature) or imaged regions (e.g., near grains boundaries vs. inside grains), providing a critical tool to support real-time TEM video analysis for material property exploration. We anticipate this YOLO based analyzing system will significantly enhance the capabilities of in-situ TEM/STEM image analysis.

**Data and Code Dissemination**

All data and code files are stored in the Materials Data Facility[72,73] at DOI: 10.18126/n9dj-5mk0. They are described in detail below.

1. Raw Data: In the folder Raw_Data, we provide the original TIFF format video and the converted 1176 JPG images of each frame and the cropped center region of interest 1176 JPG images.
2. Labeled Data: In the folder Training_and_Testing_Dataset, we provide the labeled data and the data is already put into the TRAIN folder and TEST folder. One needs to put the full path to these directories in the YOLO labeling file (called "train.txt" in our codes).
3. Code: In the code folder we provide the codes. Specifically, we provide all the codes we used in organized into Test, Train, and Trackpy subfolders of the Code folder, based on their respective applications.
4. Fitted Defects Contour Results: We provided the fitted defects of original size videos and cropped region of interest as MP4 videos in the FittedDefects_video folder.

5. Plotting Figures and Data: In the folder YOLO_Figures, we provided all the scripts and data we used to plot figures shown in this paper, and subfolders are named by the index of figures.

We also provide all codes in user-friendly IPython notebooks through GitHub at https://github.com/uw-cmg/DefectSTEMVideoAnalysis.

**Competing Interests**

There are no competing interests in relation to the work described.

**Acknowledgments**

We would like to thank the Wisconsin Applied Computing Center (WACC) for providing access to the CPU/GPU cluster, Euler. And special thanks to Colin Vanden Heuvel for helping us use GPUs and install the required software.

**Funding**[1]

The research was sponsored by the Department of Energy (DOE) Office of Nuclear Energy, Advanced Fuel Campaign of the Nuclear Technology Research and Development program (formerly the Fuel Cycle R&D program). Support for D. M. was provided by the National Science Foundation Cyberinfrastructure for Sustained Scientific Innovation (CSSI) program, award No. 1931298. Support for select undergraduate participants over some periods provided by the NSF University of Wisconsin-Madison Materials Research Science and Engineering Center (DMR 1720415) and the Schmidt Foundation. Support for the drafting phase of this work for K.G.F. was provided under a Nuclear Regulatory Commission (NRC) faculty development grant.

---

[1] Notice: This manuscript has been authored by UT-Battelle, LLC, under contract DE-AC05-00OR22725 with the US Department of Energy (DOE). The US government retains and the publisher, by accepting the article for publication, acknowledges that the US government retains a nonexclusive, paid-up, irrevocable, worldwide license to publish or reproduce the published form of this manuscript, or allow others to do so, for US government purposes. DOE will provide public access to these results of federally sponsored research in accordance with the DOE Public Access Plan (http://energy.gov/downloads/doe-public-access-plan).